\documentclass{article}

\usepackage{microtype}
\usepackage{hyperref}
\usepackage{url}
\usepackage{booktabs}
\usepackage{array}
 \usepackage{graphicx} 
 \usepackage{tabularx}
 \usepackage{multirow} 
 
\usepackage{microtype}
\usepackage{graphicx}
\usepackage{subcaption}
\usepackage{booktabs} 

\usepackage{hyperref}

\usepackage[preprint]{icml2026}

\usepackage{amsmath}
\usepackage{amssymb}
\usepackage{mathtools}
\usepackage{amsthm}
\usepackage[most]{tcolorbox}
\usepackage{lineno}
\newtcolorbox[auto counter,number within=section]{promptbox}[2][]{%
  colback=gray!5!white,
  colframe=gray!75!black,
  fonttitle=\bfseries,
  enhanced,
  breakable,
  title=Prompt~\thetcbcounter: #2,
  #1}
\definecolor{darkblue}{rgb}{0, 0, 0.5}
\hypersetup{colorlinks=true, citecolor=darkblue, linkcolor=darkblue, urlcolor=darkblue}

\usepackage[capitalize,noabbrev]{cleveref}

\theoremstyle{plain}

\theoremstyle{definition}

\theoremstyle{remark}

\usepackage[textsize=tiny]{todonotes}

\icmltitlerunning{Knowing but Not Showing: LLMs Recognize Ambiguity but Rarely Ask Clarifying Questions}

\begin{document}

\twocolumn[
  \icmltitle{Knowing but Not Showing: LLMs Recognize Ambiguity but Rarely Ask Clarifying Questions}




\begin{icmlauthorlist}
    \icmlauthor{Jinyan Su}{a}
    \icmlauthor{Claire Cardie}{a}

  \end{icmlauthorlist}

  \icmlaffiliation{a}{Cornell University }
  \icmlcorrespondingauthor{Jinyan Su}{js3673@cornell.edu}

  \icmlkeywords{Machine Learning, ICML}

  \vskip 0.3in
]



\printAffiliationsAndNotice{}  

\begin{abstract}
  User queries are often underspecified and may admit multiple valid interpretations. Rather than silently making assumptions about the user’s intent, a helpful assistant should surface such ambiguity by asking a clarifying question. Doing so requires two abilities: recognizing that a query is ambiguous, and acting on that recognition by seeking clarification instead of answering directly. To study these abilities, we evaluate models on ambiguous, unambiguous, and disambiguated questions in three settings: standard question answering, explicit ambiguity judgment, and behavioral analysis, where a judge model classifies responses as direct answers, refusals, or clarifying questions. We find a clear gap between recognition and behavior: models often identify ambiguity when explicitly asked to judge it, yet in the QA setting they overwhelmingly default to direct answers. Retrieved context further widens this gap by improving answerability while making models even less likely to ask clarifying questions.
\end{abstract}

\section{Introduction}
Large language models (LLMs) are now used at scale to answer questions, provide advice, and support decision making for a wide range of users \citep{zhang2020dialogpt, xu2023baize, achiam2023gpt, team2023gemini}. In these interactions, user queries are often not fully specified: they may leave out crucial details, rely on unstated context, or be compatible with multiple plausible interpretations \cite{zhang2024clamber}. For such ambiguous queries, simply producing a single “best guess” answer forces the model to make hidden assumptions about the user’s intent, which can misrepresent the underlying task and propagate errors into downstream decisions. Instead, an effective assistant should actively seek missing information. To do so, the model first needs ambiguity awareness: the ability to recognize that a query is ambiguous or incomplete and therefore unsafe to answer directly. It should then convert this awareness into clarification behavior, by making its uncertainty visible, either by asking a clarifying question or by explicitly refusing to answer unless more information is provided, rather than silently committing to one assumed interpretation.
\begin{table*}[h]
\centering
\small
\begin{tabularx}{\textwidth}{l X X}
\toprule
\textbf{Category} & \textbf{Concise Description} & \textbf{Example} \\
\midrule

\textbf{Temporal} &
Unspecified Time &
“Who was the president?”
(Depends entirely on the \textbf{year}) \\
\midrule

\textbf{Identity} &
Multiple entities share the same name. &
``Who is Ben Stone?'' (Several unrelated characters share this name.) \\
\midrule

\textbf{Version} &
Multiple instantiated forms of the same entity &
``Who played Annie?'' (Different actresses in the 1982, 1999, and 2014 versions.) \\
\midrule

\textbf{Scope} &
Ambiguity involving multiple levels of granularity within the same domain or entity. &
``Where did fighting happen in WWI?'' (Europe? Western Front? A single battle?) \\
\midrule

\textbf{Semantic} &
Language allows multiple interpretations &
``What is the flower of the dead in Spanish?'' (Literal translation vs.\ cultural symbol.) \\
\midrule

\textbf{Locale} &
Unspecified geographic region &
``When does MasterChef Junior start in 2018?'' (Different air dates in US vs.\ UK.) \\
\midrule

\textbf{Other} &
- &
- \\
\bottomrule
\end{tabularx}
\caption{Concise descriptions and representative examples for the ambiguity categories.}
\label{tab:ambiguity-examples}
\end{table*}

In this paper, we investigate ambiguity awareness and clarification behavior in LLMs. To probe ambiguity awareness, we ask LLMs to make explicit judgments about the query itself with or without retrieved context: decide whether a question is ambiguous and assign it to a category from our ambiguity taxonomy. On the behavior side, we examine whether this awareness carries over when model are directly prompted with the query: given the same questions, does the model actually act on its ambiguity judgments by asking for clarification or indicating that more information is needed.
\begin{figure*}[h]
    \centering
\includegraphics[width=\linewidth]{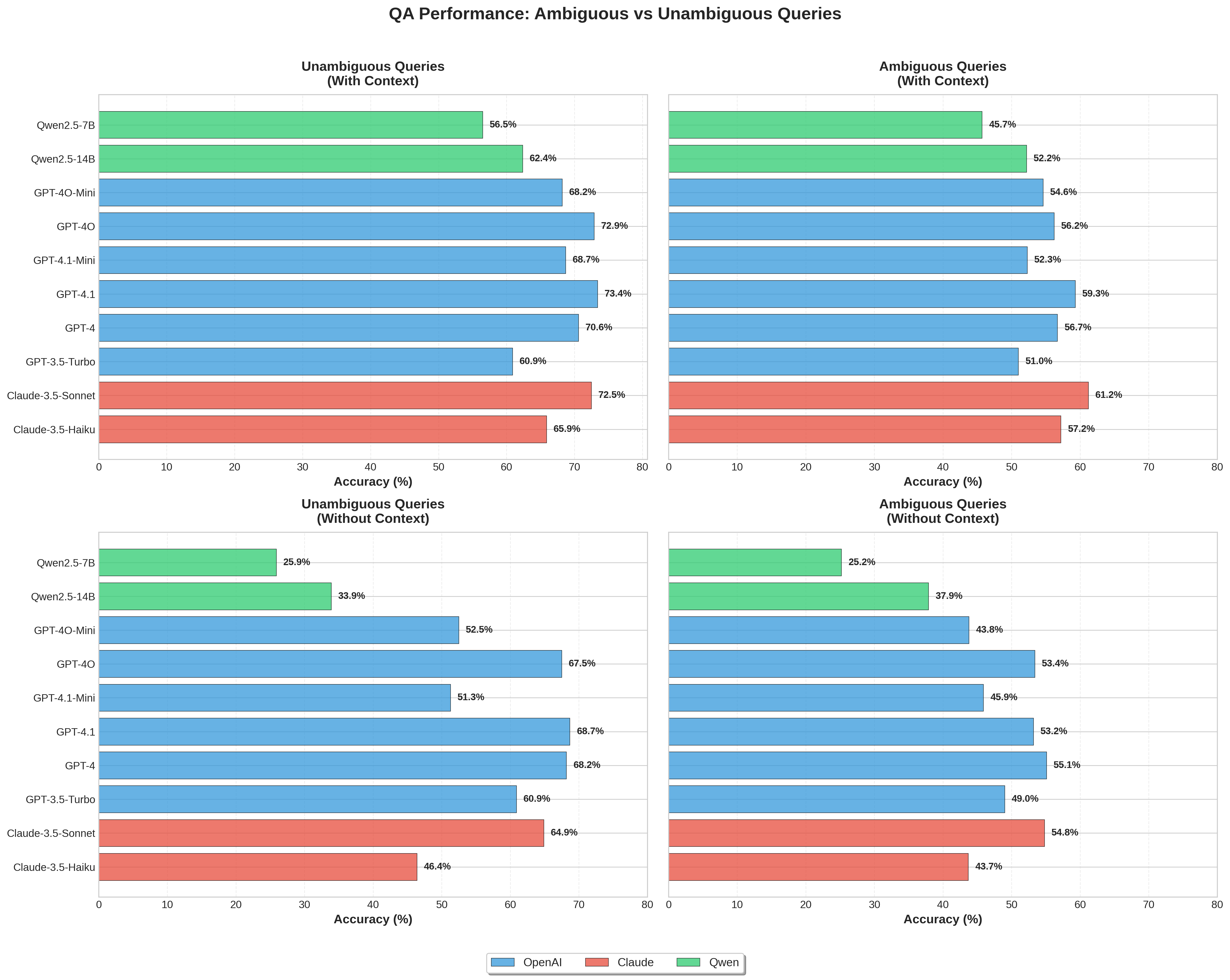}
    \caption{Overall performance in QA setting for unambiguous (left column) and ambiguous queries (right column) and also with (top row) and without (bottom row) retrieved contexts.  }
    \label{fig: qa overall performance}
\end{figure*}

We find three consistent effects. First, when the model is directly prompted with the query, the presence of retrieved context substantially improves QA accuracy across all three question categories (ambiguous, unambiguous, and disambiguated). Second, when explicitly asked to judge ambiguity, models exhibit non-trivial ambiguity awareness: they can often recognize that a query is ambiguous, and additional context sometimes helps these judgments, though the gains are modest and less uniform than the more reliable improvements observed for QA accuracy. Nevertheless, this awareness remains largely latent: even when models recognize that a query is ambiguous in this explicit judgment setting, they do not reveal that uncertainty unless they are asked to. Third, the presence of retrieved context makes models less likely to ask clarifying questions or indicate that more information is needed, regardless of whether the question itself is ambiguous or not. 

Taken together, these findings instantiate a “knowing but not showing” pattern: models can detect ambiguity to some extent, but this awareness does not reliably surface in their default answering behavior, and is further suppressed when context is provided. From a reinforcement learning perspective \cite{ouyang2022training}, current training pipelines incentivize LLMs to reveal whatever yields high reward: producing responses that human annotators rate as helpful and useful. When expressing ambiguity awareness is not aligned with high reward, models learn to hide that awareness in their default behavior.This has two important implications. First, to make models more truthful and reliable under ambiguity, we need training objectives that positively reward expressing uncertainty, asking clarification questions, and explicitly acknowledging under-specification, rather than only encouraging answer accuracy. Second, when probing what LLMs “know” or estimating their capabilities, we must be careful about how we formulate the task: the very same model can display quite different levels of ambiguity awareness depending on whether it is prompted to answer the question or to analyze the question itself. Our study makes this discrepancy visible and points toward training and evaluation schemes that better align a model’s internal awareness with its outward behavior. Although we focus on ambiguity awareness, the same reasoning applies more broadly to other model behaviors we care about, wherever truthfully expressing what the model knows is not perfectly aligned with reward. 

\section{Related Work}
\begin{figure*}[h]
    \centering
\includegraphics[width=\linewidth]{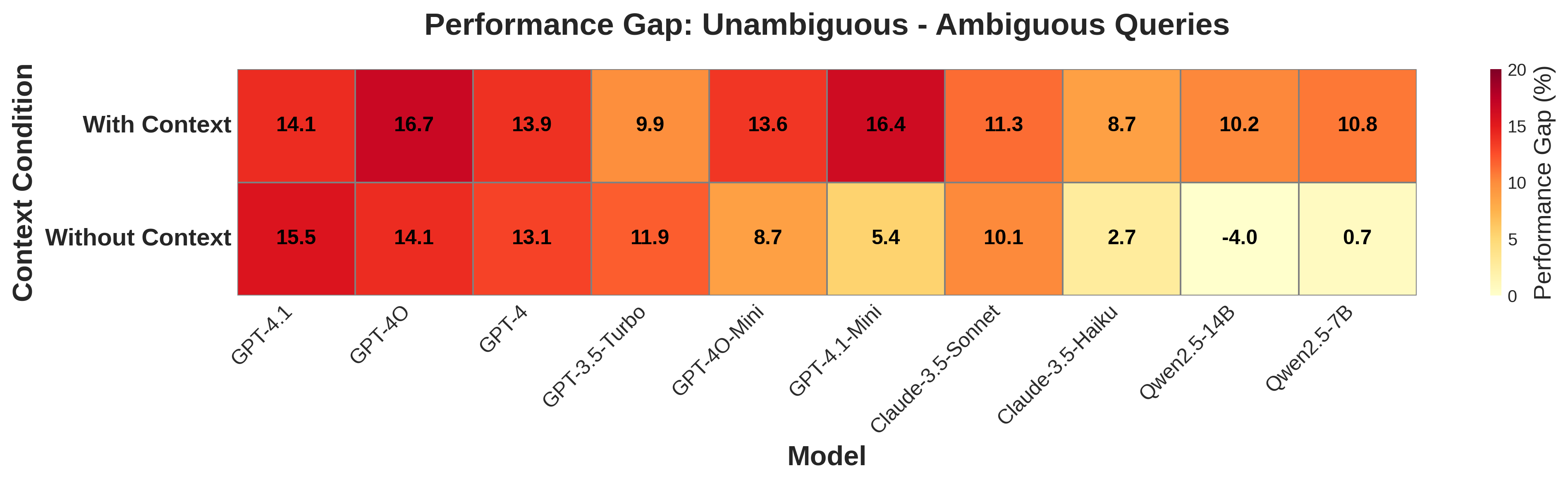}
    \caption{Performance gap of unambiguous query and ambiguous query in QA setting: model generally have a better performance on unambiguous questions with or without context.  }
    \label{fig: ambig vs unambig}
\end{figure*}
\begin{figure*}[h]
    \centering
\includegraphics[width=\linewidth]{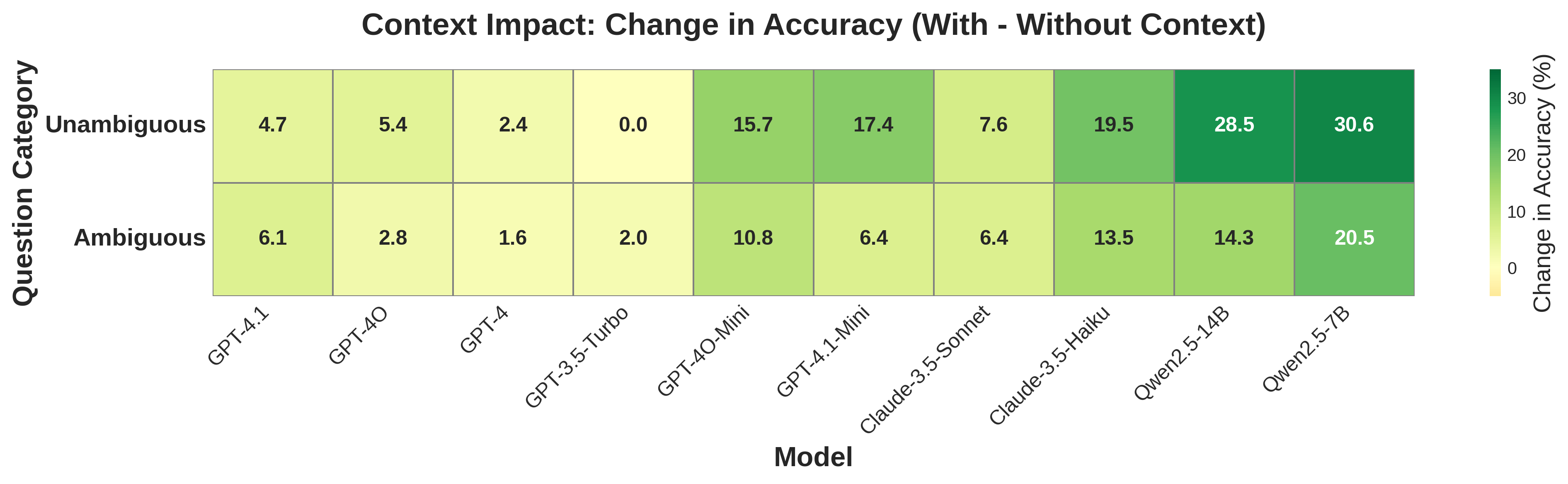}
    \caption{The impact on the performance with and without retrieved contexts: having retrieved context improves the performance for both ambiguous and unambiguous queries. }
    \label{fig: context impact}
\end{figure*}
\paragraph{Ambiguous Question Answering}
Early work has highlighted the prevalence of ambiguity in open-domain QA. \cite{min2020ambigqa} introduce AmbigQA, showing that over half of the questions in Natural Questions are underspecified and have multiple plausible answers. Follow-up studies confirm that even state-of-the-art models struggle on such ambiguous queries. For instance, \cite{wildenburg2024pre} and \cite{liu2023we} find that LLMs often underperform when questions admit multiple interpretations. \cite{stelmakh2022asqa} present ASQA, extending AmbigQA by providing each ambiguous question with a disambiguating context and long-form answers. In the conversational domain, \cite{guo2021abg} presented a dataset for detecting ambiguity in multi-turn QA and generating clarifying questions. Other datasets like CambigNQ \cite{lee2023asking} target single-turn ambiguous questions, while multi-hop ambiguity has been addressed by MIRAGE \cite{park2025mirage} for questions requiring reasoning across multiple documents. 
\paragraph{Clarifying Questions}
One prominent approach to handle ambiguity is to ask clarifying questions. In information-seeking dialogues, researchers have built datasets where systems must pose a follow-up question to clarify user intent. \cite{aliannejadi2021building} extend the Qulac dataset \cite{aliannejadi2019asking} with ClariQ \cite{aliannejadi2020convai3}, crowdsourcing single-turn conversational queries along with appropriate clarifying questions and answers. Similarly, \cite{kumar2020clarq} construct ClarQ, a clarifying question dataset derived from StackExchange posts in a QA setting. Subsequent research has treated clarifying question generation as a learning problem. For example, \cite{shridhar2023distilling} train models to generate a clarification question for ambiguous inputs and use those questions to improve downstream answering via knowledge distillation. Despite progress in supervised settings, off-the-shelf LLMs rarely ask for clarification by default.  \cite{kuhn2022clam} showed that an LLM can be prompted to either answer or ask a follow-up while \cite{deng2023prompting} find that chat-oriented LLMs like ChatGPT often fail to ask clarifying questions for ambiguous queries unless explicitly prompted. Altogether, these prior work suggests that training and prompting methods can encourage LLM to proactively resolve ambiguity, though this behavior is not yet inherent in most models’ default responses.
\paragraph{Abstention and Uncertainty}
Another line of research addresses ambiguity by allowing the model to abstain or defer answering when uncertain \cite{cole2023selectively, kamath2020selective, kadavath2022language}. \cite{shi2025ambiguity} prompt an LLM to generate multiple answers to a question, and then analyze them to infer the ambiguity. Benchmarks such as CoCoNot \cite{brahman2024art} and AbstentionBench \cite{kirichenko2025abstentionbench} contain questions that are underspecified, and require LLMs to reason about uncertainty and selectively abstain. 
\begin{figure*}[h]
    \centering
\includegraphics[width=\linewidth]{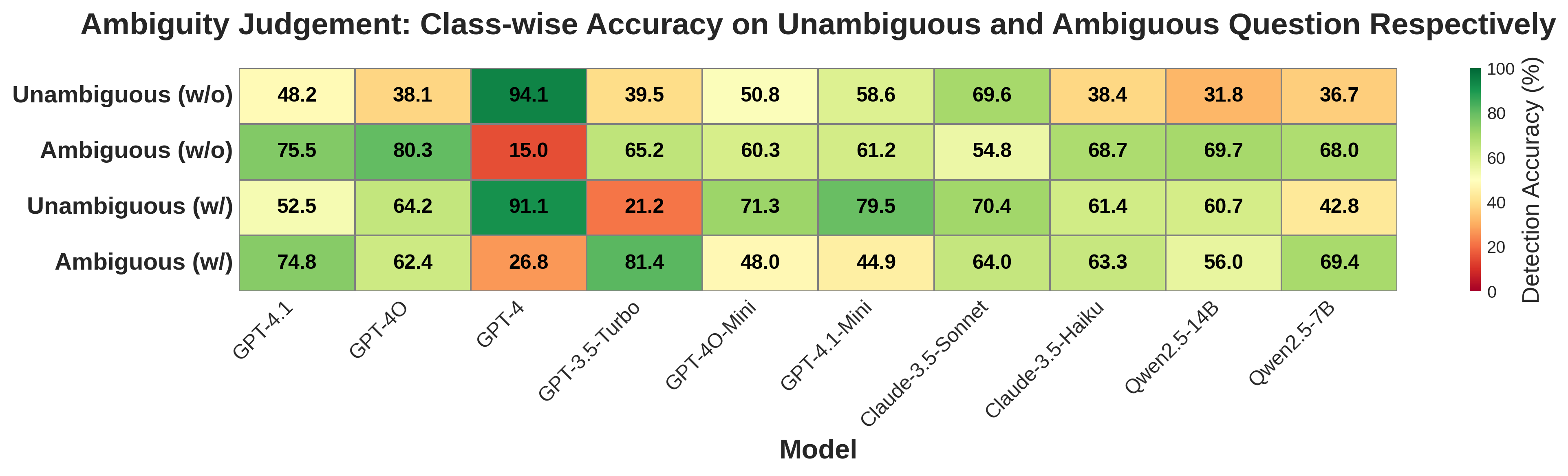}
    \caption{Classification Accuracy on ambiguous and unambiguous class respectively. (with or without given the retrieved contexts).}
    \label{fig: ambiguity judgement heatmap}
\end{figure*}
\begin{figure*}[h]
    \centering
\includegraphics[width=\linewidth]{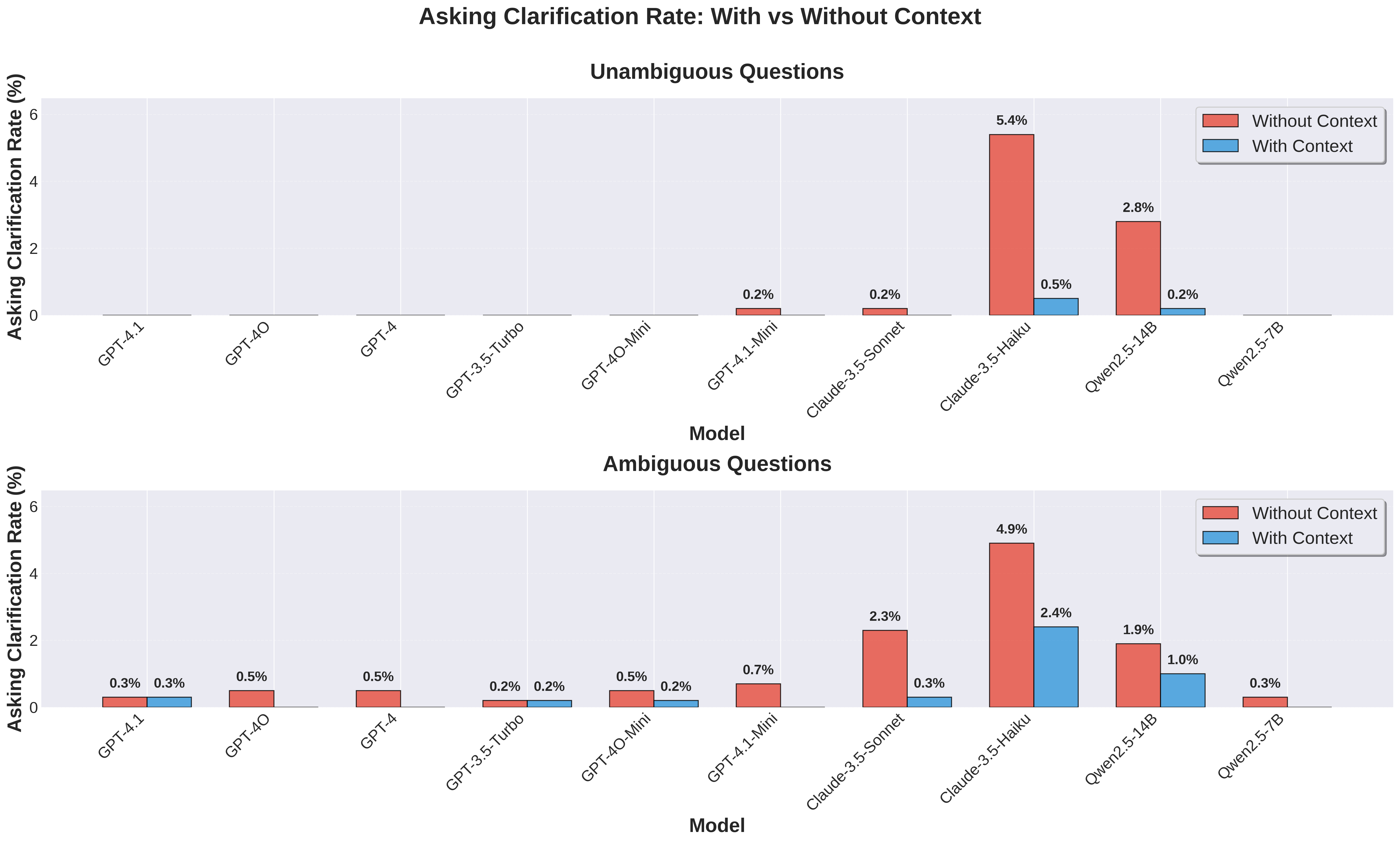}
    \caption{Asking clarification rate with and without retrieved
context. For each model (x-axis), we plot the percentage of responses that
contain any clarifying question, separately for unambiguous (top panel) and
ambiguous (bottom panel) questions; red bars correspond to runs without
context and blue bars to runs with context. }
    \label{fig:asking-clarification-comparison}
\end{figure*}
\section{Experimental Setup}
We study these questions on AmbigQA \citep{min2020ambigqa}, a dataset containing ambiguous questions, unambiguous questions, and disambiguated rewrites of the ambiguous ones. On this dataset, we conduct three complementary evaluations. First, we measure QA accuracy on ambiguous, unambiguous, and disambiguated questions, both with and without retrieved context, to quantify how model performance differs across these categories and how the presence of context affects it. Second, we probe ambiguity awareness by asking models to decide whether each question is ambiguous, again, both with and without context. Third, we analyze clarification behavior by examining, for each question, how often the model asks a clarifying question, directly answers, or explicitly refuses and indicates that more information is needed.
\subsection{Dataset}
We conduct all experiments on AmbigQA \citep{min2020ambigqa}, a question answering dataset built from open domain version of Natural Questions \cite{kwiatkowski2019natural} that identifies 
all the possible answers to an open domain question, along with the disambiguated questions. Each example consists of a question, one or more answer strings, and a set of Wikipedia passages retrieved for that question, i.e., context. For questions that are ambiguous, the dataset contains also the disambiguated version of the original ambiguous questions, as well as their respective answers. Crucially for our purposes, the dataset contains:
\begin{itemize}
    \item \textbf{Ambiguous questions}, where the original user query admits multiple plausible interpretations and is annotated with multiple disambiguated question--answer pairs.
    \item \textbf{Unambiguous questions}, which have a single clear interpretation based on the retrieved context.
    \item \textbf{Disambiguated rewrites}, where each ambiguous question is rewritten into several more specific questions, each corresponding to one underlying interpretation and answer.
\end{itemize}
 In our experiments, we randomly sample 1000 data to do all the analysis. Within this sampled set, 425 items are labeled unambiguous and 575 are labeled ambiguous. For the 575 ambiguous items, it can be disambiguated into total of 2,460 sub-questions (an average of \(\sim\)4.3 per ambiguous question). We summarize the dataset statistics and category description in Table \ref{tab:dataset_stats}.

\subsection{Task Formulation}
We evaluate each model in two explicit task settings and perform an additional behavioral analysis.

\paragraph{Question Answering.}
In the QA setting, the model is prompted with the original question from the dataset, either with or without retrieved context, and is asked to answer it directly. We then measure the correctness of the model's response against the reference answers.

\paragraph{Ambiguity Judgment.}
In the ambiguity-judgment setting, the model is asked to determine whether a question is ambiguous or unambiguous, making this setting a binary classification task. The exact prompt is shown in Prompt~\ref{box:ambiguity-detection}. This setting is designed to probe the model's explicit ambiguity awareness.
\begin{figure*}[h]
    \centering
\includegraphics[width=\linewidth]{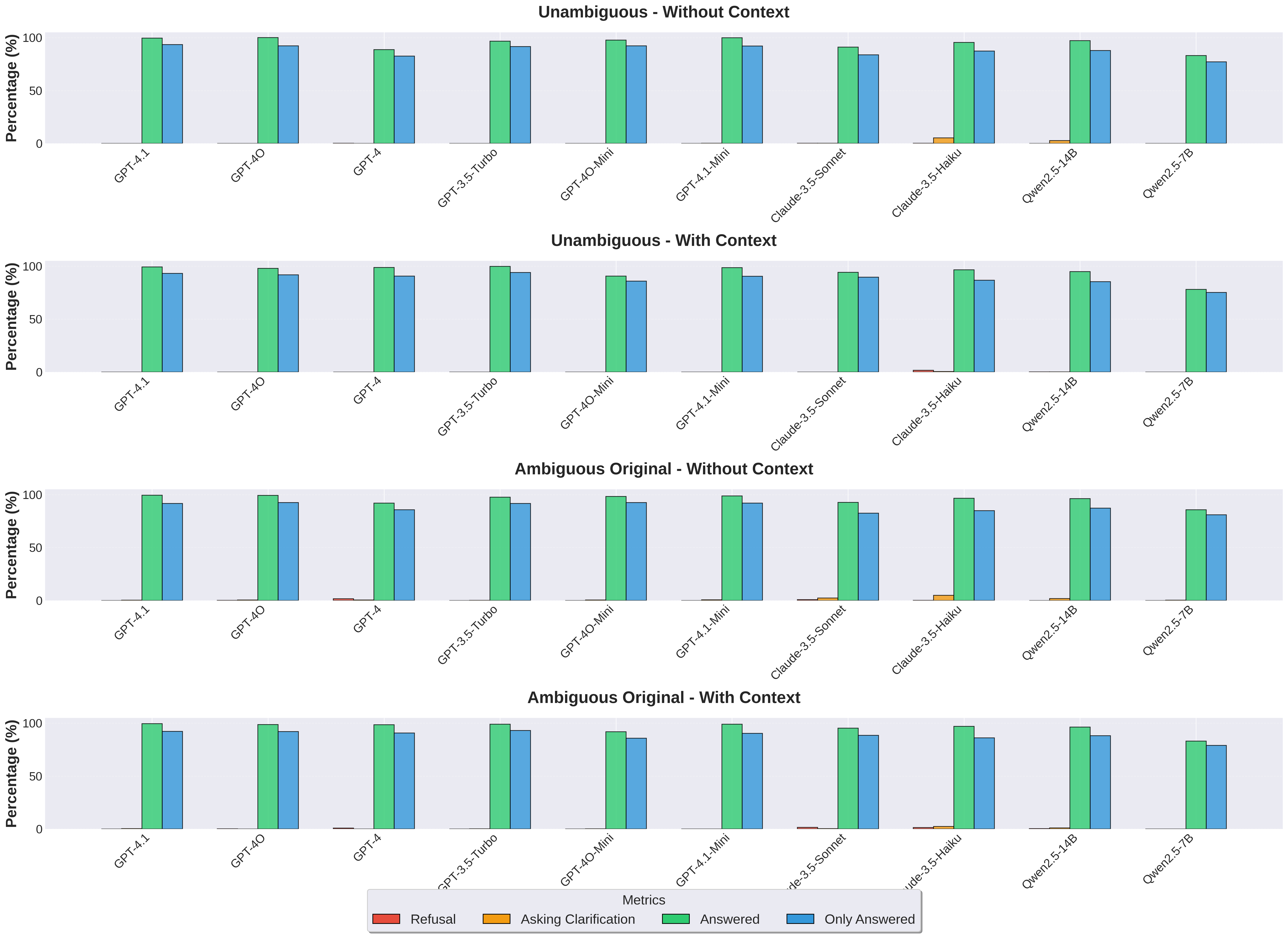}
    \caption{Behavioral breakdown of model responses. For each model, we plot the proportion of questions whose response (i) contains an explicit refusal, (ii) asks a clarifying question, (iii) contains any answer, and (iv) is a pure answer without refusal or clarification. Results are shown separately for unambiguous vs. ambiguous questions and with vs. without retrieved context.}
    \label{fig: behavioral analysis rate}
\end{figure*}
\paragraph{Behavioral Analysis.}
In addition to task accuracy, we analyze the behavioral patterns of model outputs in both settings. In the QA setting, we classify each response as a direct answer, a clarifying question, or an explicit refusal / indication that more information is needed. In the ambiguity-judgment setting, we analyze which ambiguity categories the model assigns, how these distributions vary across models, and how they compare with human annotations.
\subsection{Model and Evaluation Metrics}
We evaluated on 10 models across 3 model families, among which, 6 models from OpenAI family (GPT-4.1, GPT-4O, GPT-4, GPT 3.5-Trubo, GPT-4O-Mini, GPT-4.1-Mini), 2 models from Claude family (Claude-3.5-Sonnet and Claude-3.5-Haiku) and 2 models from Qwen family(Qwen2.5-14B and Qwen2.5-7B). In evaluation, we report accuracy as the main metrics, while the F1, Precision, Recall are reported in the Appendix for reference. 

\section{Results}
\subsection{QA setting}\label{sec: QA setting} In Figure \ref{fig: qa overall performance}, we show the performance under the QA setting for ambiguous questions and unambiguous questions, either with and without retrieved context.

From Figure \ref{fig: qa overall performance}, we can observe that (1) Comparing the left column, i.e., unambiguous queries and the right column, i.e., ambiguous queries, ambiguous queries are harder to answer correctly. For every model and both settings (with and without context), accuracy on ambiguous questions is noticeably lower than on unambiguous ones (roughly a 10–15 point drop for the stronger models). We can also notice that having retrieved context amplified the gap between ambiguous and unambiguous queries. (See Figure \ref{fig: ambig vs unambig}). For example, before for gpo-4o-mini, the performance gap of unambiguous question and ambiguous question is 5.4\% without the retrieved context, which has changed to 16.4 when the context is presented. A similar trend for Claude and Qwen family.   (2) 
Adding retrieved context boosts QA accuracy for both ambiguous and unambiguous questions. On average, unambiguous questions go from ~54\% to ~67\%, while ambiguous questions go from ~46\% to ~55\%. We can also notice that smaller and weaker models (e.g., Qwen 7B/14B, Haiku, minis) gain 15–30 points with context, while frontier models gain a more modest 2–8 points. However, with retrieval, ambiguous questions still lag behind unambiguous ones: top models hit ~70\%+ on unambiguous queries with retrieved context but only ~60\% on ambiguous with context. We can also notice that ambiguous query gains more when presented with the contexts (See Figure \ref{fig: context impact}), potentially because unambiguous questions benefit more from the extra passages while inherently ambiguous questions remain limited by missing intent rather than missing knowledge.(3) Though the ambiguity of the question and the existence of the Comparing across different models, we found that the model ranking is relatively stable. GPT-4.1 and 4o and Claude Sonnet are consistently at the top, Qwen-2.5 7B at the bottom, across all four panels. 

\subsection{Ambiguity-judgment setting}
In Section~\ref{sec: QA setting}, we considered the QA setting, where models are directly asked to answer each question. Here, we instead reformulate the task as a binary classification problem and ask the model to judge whether a question is ambiguous or not, using the prompt in Prompt~\ref{box:ambiguity-detection}. We evaluate class-wise accuracy on the gold ambiguous vs.\ unambiguous labels, with and without the same retrieved context similar to the QA setting.

Figure~\ref{fig: ambiguity judgement heatmap} shows the resulting accuracies. Most models achieve reasonably high accuracy on the ambiguous class (often 60--80\%), but are noticeably worse at recognizing unambiguous questions, especially without context (top row). This asymmetry suggests that many models are biased toward over-predicting ambiguity, achieving higher recall on ambiguous questions at the expense of accuracy on unambiguous ones.

Retrieved context does not improve recognition of ambiguous questions. Instead, its main effect is to improve performance on the unambiguous class, while accuracy on ambiguous questions remains largely flat and even declines for some models. This pattern suggests that models are not reliably using the retrieved passages to reason about whether the original query still admits multiple interpretations. Rather, the presence of supporting passages seems to act mainly as a cue that the question is answerable, making models more likely to classify it as unambiguous. If models were genuinely reasoning about ambiguity conditional on context, we would expect context to help them identify the multiple plausible interpretations of an underspecified query, not simply push them toward the unambiguous label.

GPT-4 provides an extreme example of this pattern. It achieves very high accuracy on unambiguous questions (around 90\% in both settings), but performs poorly on the ambiguous class (15--27\%). This large disparity indicates a particularly strong bias toward treating questions as unambiguous.

\subsection{Behavioral Analysis}
In this section, we move beyond task accuracy and analyze the behavior of the models. We look deeper at the free-form responses in the QA setting and measure how often they (i) refuse to answer, (ii) provide an answer, and (iii) ask a clarifying question. Because responses can contain both an answer and a clarification request, these categories are not mutually exclusive. We use \texttt{gpt-5-nano} as a judge model, prompted as in Prompt~\ref{box:judge-classification}, to label each response with these behavioral categories.

Figure~\ref{fig: behavioral analysis rate} shows the behavioral breakdown of
model outputs in the QA setting. Across all models and conditions, the
dominant behavior is to give a direct answer: answer rates are typically
above 95\%, and 80--95\% of responses are judged as \emph{only answered}
(i.e., an answer with no refusal and no clarifying question). Refusals are
almost never observed, and clarifying questions are similarly rare. Combined with the
ambiguity-judgment results, this reinforces our “knowing but not showing"
pattern: models can recognize that a query is ambiguous when explicitly
asked to judge it, but in the QA setting they almost always commit to an
answer rather than expressing that ambiguity through clarification or
refusal.

Figure~\ref{fig:asking-clarification-comparison} zooms in on the
clarification rate with and without retrieved context. Providing context slightly increases the
share of pure answers and further suppresses the already tiny rates of
clarification. Compared to other models, the Claude family is the most
likely to ask clarifying questions (up to about 5\% on ambiguous questions
without context), with Qwen2.5-14B also showing a small but visible
increase. However, even for these models, adding context consistently
reduces the clarification rate, suggesting that once supporting
passages are present, the model tends to treat the query as effectively
unambiguous and is more likely to commit to an answer. Interestingly, this suppression is stronger for
unambiguous questions than for ambiguous ones: once context is present,
most models almost never ask for clarification on unambiguous queries, and
only slightly reduce their already low clarification rate on ambiguous
queries. 

We next ask what kinds of ambiguity different models rely on when they judge a query as ambiguous. We manually define a taxonomy of six ambiguity types—Temporal, Identity, Version, Scope, Semantic, and Locale—and group all remaining cases into an ``Other'' category, for a total of seven categories. Descriptions and examples for each type are shown in Table~\ref{tab:ambiguity-examples}.

For each model, we then ask which kinds of ambiguity it relies on.
We restrict attention to ambiguous questions that the model correctly
labels as ambiguous in the ambiguity-judgment setting, and use
\texttt{gpt-5} as a judge model (Prompt~\ref{box:ambiguity-type-classification})
to map each question and explanation to one or more ambiguity types from
our taxonomy. This yields an empirical distribution over ambiguity
categories for each model, which we compare against the
human-derived distribution.

\begin{figure*}[h]
    \centering
    \includegraphics[width=\linewidth]{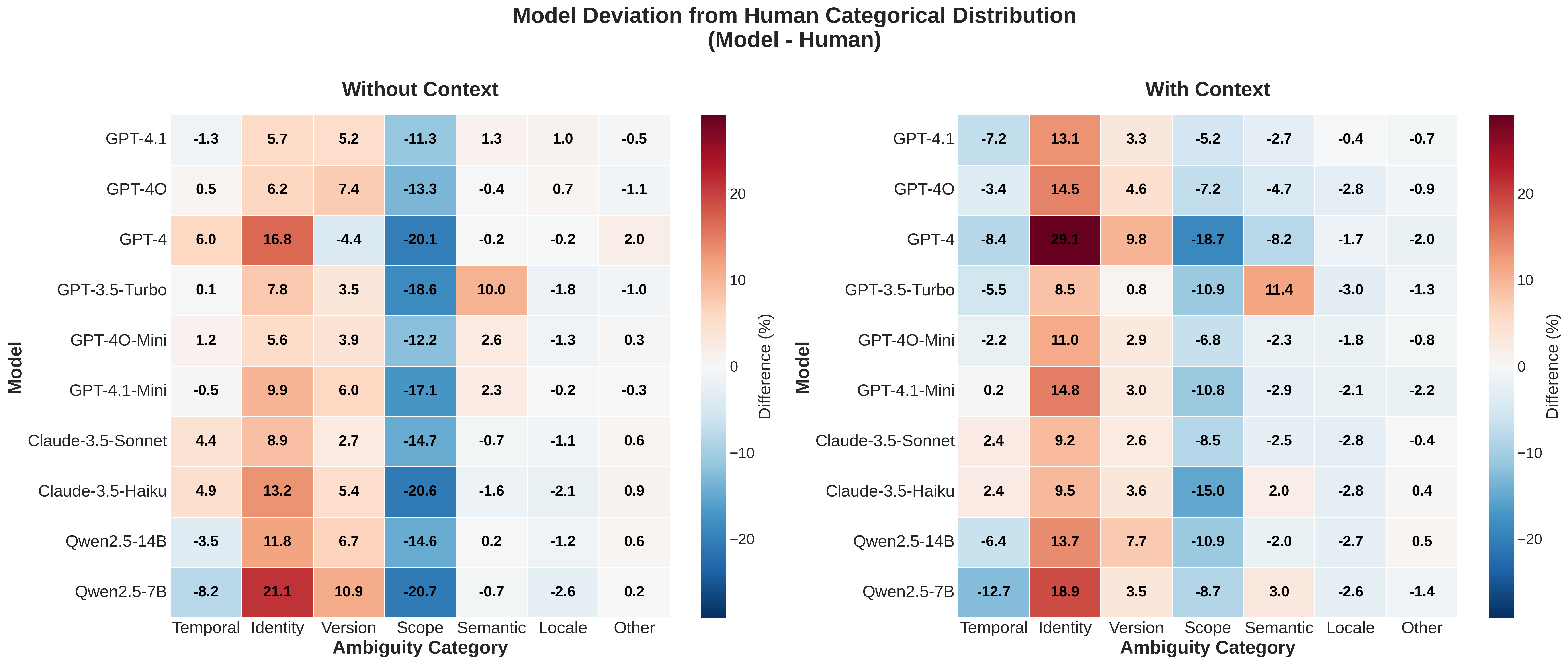}
    \caption{
    Deviation of model ambiguity-type distributions from the human
    distribution (Model -- Human, in percentage points) for ambiguous
    questions judged correctly as ambiguous. }
    \label{fig:deviation}
\end{figure*}
Figure~\ref{fig:deviation} shows, for each model, how its ambiguity-type
distribution differs from the human distribution (Model -- Human, in
percentage points). Across almost all
models and both context conditions, \emph{Identity} and, to a lesser
extent, \emph{Version} ambiguities are consistently \emph{over}-assigned
(relative to humans), while \emph{Scope} ambiguities are strongly
\emph{under}-assigned. Humans frequently explain ambiguity in terms of
missing scope information. GPT-4 amplifies this
trend most strongly, with a particularly large over-use of the Identity
category.

\section{Discussion and Future Work}
Our results point to several promising directions for future work. First, as highlighted throughout the paper, there is a misalignment between current reward signals and model faithfulness: models often know that a query is ambiguous, but their default behavior is to act as if it were fully specified. A natural next step is to design improved reward functions and training procedures that explicitly incentivize models to reveal what they know.

Second, we observe a striking asymmetry in how models use retrieved context. In the QA setting, context reliably improves performance, but in the ambiguity-judgment setting, the same context provides little benefit and can even make models less likely to flag questions as ambiguous. This suggests that the model are using superficial signals when deciding if a query is  ambiguous rather than reasoning over the given contexts. Improving models’ ability to reason over retrieved context is an important direction for enhancing their ambiguity-awareness.

\clearpage

\bibliography{example_paper}
\bibliographystyle{icml2026}

\newpage
\clearpage
\appendix
\onecolumn
\section{Dataset Statistics for evaluation}
\begin{table*}[h]
\centering
\caption{Data description and statistics used for evaluation.}
\label{tab:dataset_stats}
\begin{tabular}{l c >{\raggedright\arraybackslash}m{7cm}}
\toprule
\textbf{Question Category}  & \textbf{\# Questions} & \textbf{Description} \\
\midrule
Unambiguous & 425 &
Questions that have a single interpretation. \\\hline
Ambiguous & 575 &
Questions that have multiple interpretations. \\\hline
Disambiguated sub-questions & 2{,}460 &
Rewrites that isolate a single interpretation of an ambiguous question
(about 4.3 per ambiguous question). \\
\bottomrule
\end{tabular}
\end{table*}
\section{Prompts} \label{app: prompt}
Prompt used for judging whether the LLM's answer is refusal, asking for clarifying questions, answering or only answering.

\begin{promptbox}[label=box:judge-classification]{Judge Classification Prompt}
\small
\texttt{You are an expert judge evaluating LLM responses. Your task is to classify responses into four categories:}

1. \textbf{Refusal}: The response refuses to answer the question or states it cannot answer. This includes:
   - Explicit refusals ("I cannot", "I don't know", "I'm not able to", "I don't have information")
   - Safety refusals or policy-based refusals
   - Any response that indicates the model cannot or will not provide an answer
   - Mark as TRUE only if there's an explicit refusal, even if the model also provides some information

2. \textbf{Asked a Clarifying Question}: The response asks the user for clarification, more information, or disambiguation. This includes:
   - Questions asking for more context ("Could you clarify?", "What do you mean by...?", "Can you provide more details?")
   - Requests for disambiguation ("Are you asking about...?", "Did you mean...?")
   - Any response that seeks additional information before or while answering
   - Mark as TRUE if the response asks any clarifying question, even if it also provides an answer

3. \textbf{Answered}: The response provides an actual answer to the question, regardless of whether it also:
   - Asks for clarification
   - Refuses (though typically these don't co-occur)
   - Provides additional context or explanation
   - Mark as TRUE if there's any substantive answer to the question, even if the model guesses or is uncertain
   - Mark as TRUE even if the answer is provided alongside a clarifying question
   - Mark as FALSE only if the response contains no actual answer (e.g., only refusal, only asking for clarification without providing any answer)

4. \textbf{Only Answered}: The response provides an answer AND does NOT refuse AND does NOT ask for clarification. This means:
   - The model directly answered the question
   - No refusal is present
   - No clarifying question is asked
   - This is a pure, direct answer without additional requests or refusals
   - Mark as TRUE only if judge\_answered is TRUE AND judge\_refusal is FALSE AND judge\_asking\_clarify\_Q is FALSE

Important notes:
- These categories are NOT mutually exclusive:
  - A response can answer while also asking for clarification (judge\_answered=true, judge\_asking\_clarify\_Q=true, judge\_only\_answered=false)
  - A response can be just a refusal (judge\_refusal=true, judge\_answered=false)
  - A response can be just a clarifying question without an answer (judge\_asking\_clarify\_Q=true, judge\_answered=false)

- Be strict: only mark as refusal if there's an explicit refusal
- Be strict: only mark as asking for clarification if there's an explicit question or request
- Be lenient for "answered": mark as true if any answer is provided, even if uncertain or accompanied by clarification requests

Output your judgment as a JSON object with exactly this format:
\{
    "judge\_refusal": true or false,
    "judge\_asking\_clarify\_Q": true or false,
    "judge\_answered": true or false,
    "judge\_only\_answered": true or false
\}

Do not include any other text, only the JSON object.
\end{promptbox}

\begin{promptbox}[label=box:ambiguity-detection]{Ambiguity Detection Prompt}
\small
\texttt{You are an expert at analyzing question ambiguity. Your task is to determine whether a given question is ambiguous or unambiguous.}

An ambiguous question is one that:

- Can be interpreted in multiple ways

- Has multiple possible correct answers depending on interpretation

- Lacks sufficient context to determine the intended meaning

- Could refer to different entities, time periods, or contexts

An unambiguous question is one that:

- Has a clear, single interpretation

- Has a specific, well-defined answer

- Contains sufficient context to determine the intended meaning

- Cannot be reasonably interpreted in multiple ways

Please analyze the given question and respond with:

1. Your judgment: "AMBIGUOUS" or "UNAMBIGUOUS"

2. A brief explanation of why you made this judgment

3. If ambiguous, list the different possible interpretations

Format your response as:
JUDGMENT: [AMBIGUOUS/UNAMBIGUOUS]
EXPLANATION: [Your explanation]
INTERPRETATIONS: [If ambiguous, list different interpretations]
\end{promptbox}

\begin{promptbox}[label=box:ambiguity-type-classification]{Ambiguity Type Classification Prompt}
\small
You are an expert at classifying types of ambiguity in questions. Your task is to
classify the type(s) of ambiguity based on the question and the accompanying
explanation.

Here are the categories:

1.\ \textbf{Temporal} — ambiguity due to an unspecified or unclear time period,
   date, season, or release moment. The answer changes depending on \emph{when}
   the event or fact is situated.

2.\ \textbf{Identity} — ambiguity caused by multiple distinct entities (people,
   characters, organizations, places) sharing the same name. These are different
   things with the same label.

3.\ \textbf{Version} — ambiguity involving a single underlying work or product
   that exists in multiple instantiations (versions, editions, releases,
   remakes). These are different forms of the same entity.

4.\ \textbf{Scope} — ambiguity caused by not specifying which level, subset, or
   granularity of a multi-level domain or entity is intended (e.g., franchise vs.
   team, region vs. country, entire war vs. individual battles). The variation is
   structural, not geographic.

5.\ \textbf{Semantic} — ambiguity arising from the wording itself, including
   polysemy, translation ambiguity, or syntactic ambiguity. The uncertainty is
   due to language, not entities or facts.

6.\ \textbf{Locale} — ambiguity caused by an unspecified geographic region,
   country, or local convention. Different locales have different facts, naming
   systems, or release schedules. The answer changes depending on \emph{where}.

7.\ \textbf{Other} — ambiguity that does not fit the above categories, such as
   incomplete phrasing, pragmatic underspecification, or contextual references
   without clear referents.

For each case, you should:
\begin{itemize}
  \item Identify \textbf{all} applicable ambiguity types (one or more categories),
  \item List them in order of prominence or importance,
  \item Take into account what the explanation emphasizes,
  \item If only one type applies, still use the array format with one element.
\end{itemize}

Output your classification as a JSON object with this exact format:
\begin{quote}\small
\texttt{\{}\\
\quad\texttt{"categories": [<array of numbers 1-7>],}\\
\quad\texttt{"category\_names": ["<name of category 1>", "<name of category 2>", ...],}\\
\quad\texttt{"primary\_category": <number 1-7 of the most prominent type>,}\\
\quad\texttt{"reasoning": "<brief explanation of why these categories were chosen>"}\\
\texttt{\}}
\end{quote}

Example for a question with multiple ambiguity types:
\begin{quote}\small
\texttt{\{}\\
\quad\texttt{"categories": [1, 4],}\\
\quad\texttt{"category\_names": ["Temporal", "Scope"],}\\
\quad\texttt{"primary\_category": 1,}\\
\quad\texttt{"reasoning": "The question lacks temporal context (when) and also has}\\
\quad\texttt{scope ambiguity (which level or subset is intended)."}\\
\texttt{\}}
\end{quote}

Example for a question with a single ambiguity type:
\begin{quote}\small
\texttt{\{}\\
\quad\texttt{"categories": [2],}\\
\quad\texttt{"category\_names": ["Identity"],}\\
\quad\texttt{"primary\_category": 2,}\\
\quad\texttt{"reasoning": "Multiple distinct entities share the same name."}\\
\texttt{\}}
\end{quote}

Do not include any other text, only the JSON object.
\end{promptbox}

\section{Additional Metrics}
\begin{itemize}
    \item \textbf{Disambiguated (Micro-Average):} \emph{Question-level} accuracy micro-averaged over the full pool of disambiguated sub-questions derived from the 575 ambiguous originals (i.e., over all \textbf{2,460} sub-questions). This corresponds to the average accuracy over all disambiguated items.
    \item \textbf{Disambiguated (Strict):} \emph{Example-level} accuracy at the level of each original ambiguous question: counted correct only if \emph{all} of its disambiguated sub-questions are answered correctly.
    \item \textbf{Disambiguated (Lenient):} \emph{Example-level} accuracy on the ambiguous original: counted correct if, when presented with the ambiguous question, the model’s answer matches \emph{any one} of the ground-truth answers for its disambiguated sub-questions.
\end{itemize}

\begin{table*}[h]
\centering
\caption{Accuracy (\%) \emph{without} retrieved context. Parentheses show the absolute change when adding retrieved context (\emph{with}–\emph{without}). Red = improvement; green = degradation. Columns: \textbf{UnAbg} = unambiguous originals (425 Qs); \textbf{Abg} = ambiguous originals (575 Qs); \textbf{DisAbg (Micro-Average)} = micro-averaged accuracy over all 2{,}460 disambiguated sub-questions; \textbf{DisAbg (Strict)} = example-level accuracy requiring \emph{all} sub-questions for an ambiguous original to be correct; \textbf{DisAbg (Lenient)} = example-level accuracy counted correct the answered any subquestions correctly. }

\label{tab:context_comparison}
\scriptsize
\begin{tabular}{lccccc}
\toprule
& & & \multicolumn{3}{c}{\textbf{Disambiguated}}\\\cline{4-6}
\textbf{Model} & \textbf{Unambiguous} & \textbf{Ambiguous} & \textbf{Micro-Average} & \textbf{Strict} & \textbf{Lenient} \\
\midrule
GPT-4.1 & \textbf{68.7} {\textcolor{green!60!black}{\tiny (+4.7)}} & 53.2 {\textcolor{green!60!black}{\tiny (+6.1)}} & \underline{55.0 }{\textcolor{green!60!black}{\tiny (+3.9)}} & \underline{21.4 }{\textcolor{green!60!black}{\tiny (+3.3)}} & 63.5 {\textcolor{green!60!black}{\tiny (+7.3)}} \\
GPT-4O & 67.5 {\textcolor{green!60!black}{\tiny (+5.4)}} & 53.4 {\textcolor{green!60!black}{\tiny (+2.8)}} & \textbf{55.4} {\textcolor{green!60!black}{\tiny (+2.0)}} & 20.2 {\textcolor{green!60!black}{\tiny (+3.6)}} & 64.2 {\textcolor{green!60!black}{\tiny (+2.6)}} \\
GPT-4 & \underline{68.2}{\textcolor{green!60!black}{\tiny (+2.4)}} &\textbf{55.1}{\textcolor{green!60!black}{\tiny (+1.6)}} & 52.9 {\textcolor{red}{\tiny (-2.0)}} & 20.2 {\textcolor{green!60!black}{\tiny (+0.5)}} & \textbf{65.9 }{\textcolor{green!60!black}{\tiny (+1.2)}} \\
GPT-3.5-Turbo & 60.9 {\textcolor{black}{\tiny (0.0)}} & 49.0 {\textcolor{green!60!black}{\tiny (+2.0)}} & 45.5 {\textcolor{red}{\tiny (-3.1)}} & 12.9 {\textcolor{green!60!black}{\tiny (+0.8)}} & 58.6 {\textcolor{green!60!black}{\tiny (+0.5)}} \\
GPT-4O-Mini & 52.5 {\textcolor{green!60!black}{\tiny (+15.7)}} & 43.8 {\textcolor{green!60!black}{\tiny (+10.8)}} & 40.2 {\textcolor{green!60!black}{\tiny (+6.8)}} & 11.1 {\textcolor{green!60!black}{\tiny (+4.6)}} & 55.1 {\textcolor{green!60!black}{\tiny (+8.9)}} \\
GPT-4.1-Mini & 51.3 {\textcolor{green!60!black}{\tiny (+17.4)}} & 45.9 {\textcolor{green!60!black}{\tiny (+6.4)}} & 41.6 {\textcolor{green!60!black}{\tiny (+3.0)}} & 11.8 {\textcolor{green!60!black}{\tiny (+3.9)}} & 54.3 {\textcolor{green!60!black}{\tiny (+7.8)}} \\
Claude-3.5-Sonnet & 64.9 {\textcolor{green!60!black}{\tiny (+7.6)}} & \underline{54.8 }{\textcolor{green!60!black}{\tiny (+6.4)}} & 54.3 {\textcolor{red}{\tiny (-0.3)}} & \textbf{21.6 }{\textcolor{green!60!black}{\tiny (+0.8)}} & \underline{64.9} {\textcolor{green!60!black}{\tiny (+6.6)}} \\
Claude-3.5-Haiku & 46.4 {\textcolor{green!60!black}{\tiny (+19.5)}} & 43.7 {\textcolor{green!60!black}{\tiny (+13.5)}} & 36.3 {\textcolor{green!60!black}{\tiny (+7.2)}} & 10.8 {\textcolor{green!60!black}{\tiny (+3.5)}} & 55.0 {\textcolor{green!60!black}{\tiny (+13.0)}} \\
Qwen2.5-14B & 33.9 {\textcolor{green!60!black}{\tiny (+28.5)}} & 37.9 {\textcolor{green!60!black}{\tiny (+14.3)}} & 28.3 {\textcolor{green!60!black}{\tiny (+14.4)}} & 7.1 {\textcolor{green!60!black}{\tiny (+6.3)}} & 46.6 {\textcolor{green!60!black}{\tiny (+19.7)}} \\
Qwen2.5-7B & 25.9 {\textcolor{green!60!black}{\tiny (+30.6)}} & 25.2 {\textcolor{green!60!black}{\tiny (+20.5)}} & 18.5 {\textcolor{green!60!black}{\tiny (+14.8)}} & 3.3 {\textcolor{green!60!black}{\tiny (+6.1)}} & 32.2 {\textcolor{green!60!black}{\tiny (+27.6)}} \\
\bottomrule
\end{tabular}
\end{table*}

\begin{table*}[h]
\centering
\caption{Ambiguity Judgement setting: With vs. Without Context}
\label{tab:ambiguity_detection_comparison}
\scriptsize
\begin{tabular}{lccc}
\toprule
\textbf{Model} & \textbf{Unambiguous} & \textbf{Ambiguous} & \textbf{Disambiguated Rewrites} \\
\midrule
GPT-4.1 & 48.2 {\textcolor{green!60!black}{\tiny (+4.3)}} & 75.5 {\textcolor{red}{\tiny (-0.7)}} & 47.8 {\textcolor{red}{\tiny (-3.8)}} \\
GPT-4O & 38.1 {\textcolor{green!60!black}{\tiny (+26.1)}} & 80.3 {\textcolor{red}{\tiny (-17.9)}} & 34.6 {\textcolor{green!60!black}{\tiny (+11.5)}} \\
GPT-4 & 94.1 {\textcolor{red}{\tiny (-3.0)}} & 15.0 {\textcolor{green!60!black}{\tiny (+11.8)}} & 94.3 {\textcolor{red}{\tiny (-7.3)}} \\
GPT-3.5-Turbo & 39.5 {\textcolor{red}{\tiny (-18.3)}} & 65.2 {\textcolor{green!60!black}{\tiny (+16.2)}} & 44.1 {\textcolor{red}{\tiny (-31.3)}} \\
GPT-4O-Mini & 50.8 {\textcolor{green!60!black}{\tiny (+20.5)}} & 60.3 {\textcolor{red}{\tiny (-12.3)}} & 52.3 {\textcolor{red}{\tiny (-0.7)}} \\
GPT-4.1-Mini & 58.6 {\textcolor{green!60!black}{\tiny (+20.9)}} & 61.2 {\textcolor{red}{\tiny (-16.3)}} & 52.8 {\textcolor{green!60!black}{\tiny (+12.3)}} \\
Claude-3.5-Sonnet & 69.6 {\textcolor{green!60!black}{\tiny (+0.8)}} & 54.8 {\textcolor{green!60!black}{\tiny (+9.2)}} & 62.9 {\textcolor{red}{\tiny (-9.6)}} \\
Claude-3.5-Haiku & 38.4 {\textcolor{green!60!black}{\tiny (+23.0)}} & 68.7 {\textcolor{red}{\tiny (-5.4)}} & 38.5 {\textcolor{green!60!black}{\tiny (+3.5)}} \\
Qwen2.5-14B & 31.8 {\textcolor{green!60!black}{\tiny (+28.9)}} & 69.7 {\textcolor{red}{\tiny (-13.7)}} & 50.6 {\textcolor{red}{\tiny (-1.2)}} \\
Qwen2.5-7B & 36.7 {\textcolor{green!60!black}{\tiny (+6.1)}} & 68.0 {\textcolor{green!60!black}{\tiny (+1.4)}} & 54.1 {\textcolor{red}{\tiny (-22.9)}} \\
\bottomrule
\end{tabular}
\end{table*}
\begin{table*}[h]
\centering
\caption{Refusal Rate: With vs. Without Context}
\label{tab:judge_refusal_comparison}
\scriptsize
\begin{tabular}{lccc}
\toprule
\textbf{Model} & \textbf{Unambiguous} & \textbf{Ambiguous Original} & \textbf{Ambiguous Disambiguated} \\
\midrule
GPT-4.1 & 0.0 {\textcolor{black}{\tiny (0.0)}} & 0.0 {\textcolor{black}{\tiny (0.0)}} & 0.0 {\textcolor{black}{\tiny (0.0)}} \\
GPT-4O & 0.0 {\textcolor{black}{\tiny (0.0)}} & 0.2 {\textcolor{black}{\tiny (0.0)}} & 0.0 {\textcolor{black}{\tiny (0.0)}} \\
GPT-4 & 0.2 {\textcolor{green!60!black}{\tiny (-0.2)}} & 1.7 {\textcolor{green!60!black}{\tiny (-0.8)}} & 1.2 {\textcolor{green!60!black}{\tiny (-0.5)}} \\
GPT-3.5-Turbo & 0.0 {\textcolor{black}{\tiny (0.0)}} & 0.0 {\textcolor{black}{\tiny (0.0)}} & 0.0 {\textcolor{black}{\tiny (0.0)}} \\
GPT-4O-Mini & 0.0 {\textcolor{black}{\tiny (0.0)}} & 0.0 {\textcolor{black}{\tiny (0.0)}} & 0.0 {\textcolor{black}{\tiny (0.0)}} \\
GPT-4.1-Mini & 0.0 {\textcolor{black}{\tiny (0.0)}} & 0.0 {\textcolor{black}{\tiny (0.0)}} & 0.0 {\textcolor{black}{\tiny (0.0)}} \\
Claude-3.5-Sonnet & 0.2 {\textcolor{green!60!black}{\tiny (-0.2)}} & 0.9 {\textcolor{red}{\tiny (+0.7)}} & 1.0 {\textcolor{red}{\tiny (+2.1)}} \\
Claude-3.5-Haiku & 0.2 {\textcolor{red}{\tiny (+1.4)}} & 0.2 {\textcolor{red}{\tiny (+1.2)}} & 4.4 {\textcolor{red}{\tiny (+4.9)}} \\
Qwen2.5-14B & 0.0 {\textcolor{red}{\tiny (+0.2)}} & 0.0 {\textcolor{red}{\tiny (+0.3)}} & 0.0 {\textcolor{red}{\tiny (+0.2)}} \\
Qwen2.5-7B & 0.0 {\textcolor{black}{\tiny (0.0)}} & 0.0 {\textcolor{black}{\tiny (0.0)}} & 0.0 {\textcolor{red}{\tiny (+0.1)}} \\
\bottomrule
\end{tabular}
\end{table*}

\begin{table*}[h]
\centering
\caption{Asking Clarification Rate: With vs. Without Context}
\label{tab:judge_asking_clarify_comparison}
\scriptsize
\begin{tabular}{lccc}
\toprule
\textbf{Model} & \textbf{Unambiguous} & \textbf{Ambiguous Original} & \textbf{Ambiguous Disambiguated} \\
\midrule
GPT-4.1 & 0.0 {\textcolor{black}{\tiny (0.0)}} & 0.3 {\textcolor{black}{\tiny (0.0)}} & 0.0 {\textcolor{black}{\tiny (0.0)}} \\
GPT-4O & 0.0 {\textcolor{black}{\tiny (0.0)}} & 0.5 {\textcolor{green!60!black}{\tiny (-0.5)}} & 0.1 {\textcolor{green!60!black}{\tiny (-0.1)}} \\
GPT-4 & 0.0 {\textcolor{black}{\tiny (0.0)}} & 0.5 {\textcolor{green!60!black}{\tiny (-0.5)}} & 0.2 {\textcolor{green!60!black}{\tiny (-0.2)}} \\
GPT-3.5-Turbo & 0.0 {\textcolor{black}{\tiny (0.0)}} & 0.2 {\textcolor{black}{\tiny (0.0)}} & 0.0 {\textcolor{black}{\tiny (0.0)}} \\
GPT-4O-Mini & 0.0 {\textcolor{black}{\tiny (0.0)}} & 0.5 {\textcolor{green!60!black}{\tiny (-0.3)}} & 0.0 {\textcolor{black}{\tiny (0.0)}} \\
GPT-4.1-Mini & 0.2 {\textcolor{green!60!black}{\tiny (-0.2)}} & 0.7 {\textcolor{green!60!black}{\tiny (-0.7)}} & 0.1 {\textcolor{black}{\tiny (0.0)}} \\
Claude-3.5-Sonnet & 0.2 {\textcolor{green!60!black}{\tiny (-0.2)}} & 2.3 {\textcolor{green!60!black}{\tiny (-2.0)}} & 0.5 {\textcolor{green!60!black}{\tiny (-0.5)}} \\
Claude-3.5-Haiku & 5.4 {\textcolor{green!60!black}{\tiny (-4.9)}} & 4.9 {\textcolor{green!60!black}{\tiny (-2.5)}} & 5.3 {\textcolor{green!60!black}{\tiny (-3.3)}} \\
Qwen2.5-14B & 2.8 {\textcolor{green!60!black}{\tiny (-2.6)}} & 1.9 {\textcolor{green!60!black}{\tiny (-0.9)}} & 1.6 {\textcolor{green!60!black}{\tiny (-0.5)}} \\
Qwen2.5-7B & 0.0 {\textcolor{black}{\tiny (0.0)}} & 0.3 {\textcolor{green!60!black}{\tiny (-0.3)}} & 0.0 {\textcolor{black}{\tiny (0.0)}} \\
\bottomrule
\end{tabular}
\end{table*}

\begin{table*}[h]
\centering
\caption{Answered Rate: With vs. Without Context}
\label{tab:judge_answered_comparison}
\scriptsize
\begin{tabular}{lccc}
\toprule
\textbf{Model} & \textbf{Unambiguous} & \textbf{Ambiguous Original} & \textbf{Ambiguous Disambiguated} \\
\midrule
GPT-4.1 & 99.5 {\textcolor{green!60!black}{\tiny (-0.2)}} & 99.5 {\textcolor{black}{\tiny (0.0)}} & 100.0 {\textcolor{black}{\tiny (0.0)}} \\
GPT-4O & 100.0 {\textcolor{green!60!black}{\tiny (-2.1)}} & 99.3 {\textcolor{green!60!black}{\tiny (-0.7)}} & 100.0 {\textcolor{green!60!black}{\tiny (-0.1)}} \\
GPT-4 & 88.7 {\textcolor{red}{\tiny (+10.1)}} & 92.0 {\textcolor{red}{\tiny (+6.4)}} & 98.7 {\textcolor{green!60!black}{\tiny (-2.0)}} \\
GPT-3.5-Turbo & 96.7 {\textcolor{red}{\tiny (+3.1)}} & 97.6 {\textcolor{red}{\tiny (+1.4)}} & 99.8 {\textcolor{green!60!black}{\tiny (-0.2)}} \\
GPT-4O-Mini & 97.6 {\textcolor{green!60!black}{\tiny (-7.0)}} & 98.3 {\textcolor{green!60!black}{\tiny (-6.5)}} & 100.0 {\textcolor{green!60!black}{\tiny (-0.1)}} \\
GPT-4.1-Mini & 99.8 {\textcolor{green!60!black}{\tiny (-1.2)}} & 98.8 {\textcolor{red}{\tiny (+0.2)}} & 100.0 {\textcolor{green!60!black}{\tiny (-0.7)}} \\
Claude-3.5-Sonnet & 91.1 {\textcolor{red}{\tiny (+3.0)}} & 92.7 {\textcolor{red}{\tiny (+2.6)}} & 98.4 {\textcolor{green!60!black}{\tiny (-5.8)}} \\
Claude-3.5-Haiku & 95.5 {\textcolor{red}{\tiny (+1.0)}} & 96.5 {\textcolor{red}{\tiny (+0.5)}} & 90.5 {\textcolor{green!60!black}{\tiny (-6.1)}} \\
Qwen2.5-14B & 97.2 {\textcolor{green!60!black}{\tiny (-2.4)}} & 96.3 {\textcolor{black}{\tiny (0.0)}} & 99.8 {\textcolor{green!60!black}{\tiny (-0.9)}} \\
Qwen2.5-7B & 83.1 {\textcolor{green!60!black}{\tiny (-5.0)}} & 85.7 {\textcolor{green!60!black}{\tiny (-2.6)}} & 99.5 {\textcolor{green!60!black}{\tiny (-2.3)}} \\
\bottomrule
\end{tabular}
\end{table*}

\begin{table*}[h]
\centering
\caption{Only Answered Rate: With vs. Without Context}
\label{tab:judge_only_answered_comparison}
\scriptsize
\begin{tabular}{lccc}
\toprule
\textbf{Model} & \textbf{Unambiguous} & \textbf{Ambiguous Original} & \textbf{Ambiguous Disambiguated} \\
\midrule
GPT-4.1 & 93.4 {\textcolor{green!60!black}{\tiny (-0.2)}} & 91.7 {\textcolor{red}{\tiny (+0.5)}} & 93.1 {\textcolor{red}{\tiny (+0.8)}} \\
GPT-4O & 92.2 {\textcolor{green!60!black}{\tiny (-0.4)}} & 92.5 {\textcolor{green!60!black}{\tiny (-0.4)}} & 92.7 {\textcolor{red}{\tiny (+0.4)}} \\
GPT-4 & 82.6 {\textcolor{red}{\tiny (+8.0)}} & 85.7 {\textcolor{red}{\tiny (+5.0)}} & 93.4 {\textcolor{green!60!black}{\tiny (-3.6)}} \\
GPT-3.5-Turbo & 91.5 {\textcolor{red}{\tiny (+2.4)}} & 91.7 {\textcolor{red}{\tiny (+1.3)}} & 93.7 {\textcolor{green!60!black}{\tiny (-0.8)}} \\
GPT-4O-Mini & 92.2 {\textcolor{green!60!black}{\tiny (-6.3)}} & 92.5 {\textcolor{green!60!black}{\tiny (-6.8)}} & 92.8 {\textcolor{red}{\tiny (+1.1)}} \\
GPT-4.1-Mini & 92.0 {\textcolor{green!60!black}{\tiny (-1.6)}} & 92.0 {\textcolor{green!60!black}{\tiny (-1.6)}} & 93.9 {\textcolor{green!60!black}{\tiny (-2.0)}} \\
Claude-3.5-Sonnet & 83.8 {\textcolor{red}{\tiny (+5.8)}} & 82.4 {\textcolor{red}{\tiny (+6.1)}} & 92.0 {\textcolor{green!60!black}{\tiny (-6.2)}} \\
Claude-3.5-Haiku & 87.3 {\textcolor{green!60!black}{\tiny (-0.7)}} & 84.9 {\textcolor{red}{\tiny (+1.2)}} & 79.1 {\textcolor{green!60!black}{\tiny (-7.8)}} \\
Qwen2.5-14B & 87.8 {\textcolor{green!60!black}{\tiny (-2.4)}} & 87.3 {\textcolor{red}{\tiny (+0.9)}} & 89.1 {\textcolor{red}{\tiny (+0.3)}} \\
Qwen2.5-7B & 77.2 {\textcolor{green!60!black}{\tiny (-2.1)}} & 80.9 {\textcolor{green!60!black}{\tiny (-1.9)}} & 92.1 {\textcolor{green!60!black}{\tiny (-1.2)}} \\
\bottomrule
\end{tabular}
\end{table*}
\begin{table}[h]
\centering
\begin{tabular}{lcc|cc}
\toprule
\multirow{2}{*}{Model} & \multicolumn{2}{c}{Unambiguous} & \multicolumn{2}{c}{Ambiguous} \\
\cmidrule(lr){2-3} \cmidrule(lr){4-5}
 & No Context & With Context & No Context & With Context \\
\midrule
gpt-3.5-turbo & 54.4 & 59.5 & 45.7 & 49.4 \\
gpt-4 & 67.3 & 69.6 & 53.6 & 55.8 \\
gpt-4.1 & 68.7 & 74.1 & 52.0 & 60.9 \\
gpt-4.1-mini & 52.0 & 68.2 & 46.1 & 53.9 \\
gpt-4o & 68.7 & 72.0 & 53.6 & 57.7 \\
gpt-4o-mini & 54.1 & 68.7 & 47.8 & 53.0 \\
claude-sonnet-4-5-20250929 & 62.1 & 69.9 & 50.3 & 57.4 \\
claude-haiku-4-5-20251001 & 39.1 & 62.4 & 35.7 & 53.6 \\
Qwen Qwen2.5-7B-Instruct & 24.0 & 54.4 & 21.9 & 46.1 \\
Qwen Qwen2.5-14B-Instruct & 31.3 & 61.6 & 32.9 & 50.4 \\
\bottomrule
\end{tabular}
\caption{Accuracy (EM) for models with --refuse\_if\_ambiguous flag}
\label{tab:refuse_ambiguous_results}
\end{table}

\end{document}